\documentclass{IEEEconf}

\usepackage{multirow}
\usepackage{amsmath,amssymb}
\usepackage{algorithm,algcompatible}
\usepackage{float}
\usepackage[framemethod=tikz]{mdframed}
\usepackage{soul}
\usepackage{caption,multicol}
\usepackage{hyperref}
\usepackage{wrapfig}

\newcommand{\mrg}[1]{
\makeatletter\def\SOUL@hlpreamble{\setul{}{#1}\let\SOUL@stcolor\SOUL@hlcolor\SOUL@stpreamble}\makeatother}
\newcommand{\ci}{0.5}
\definecolor{AdamColor}{rgb}{\ci,1,1}
\definecolor{AdamAMSColor}{rgb}{1,\ci,1}
\definecolor{ProxColor}{rgb}{1,1,\ci}
\definecolor{ProxAdamColor}{rgb}{\ci,\ci,\ci}
\newcommand{\hladam}[1]{{\sethlcolor{AdamColor}\hl{#1}}}
\newcommand{\hlams}[1]{{\sethlcolor{AdamAMSColor}\hl{#1}}}
\newcommand{\hlprox}[1]{{\sethlcolor{ProxColor}\hl{#1}}}
\newcommand{\hlpadam}[1]{{\sethlcolor{ProxAdamColor}\hl{#1}}}

\definecolor{AvgColor}{rgb}{\ci,1,1}
\definecolor{Adm}{rgb}{1,\ci,1}
\definecolor{YogiColor}{rgb}{1,1,\ci}
\definecolor{AdaptColor}{rgb}{\ci,\ci,\ci}
\definecolor{AdagradColor}{rgb}{\ci,1,\ci}
\newcommand{\hlfav}[1]{{\sethlcolor{AvgColor}\hl{#1}}}
\newcommand{\hlfadam}[1]{{\sethlcolor{Adm}\hl{#1}}}
\newcommand{\hlfyogi}[1]{{\sethlcolor{YogiColor}\hl{#1}}}
\newcommand{\hlfadapt}[1]{{\sethlcolor{AdaptColor}\hl{#1}}}
\newcommand{\hlfada}[1]{{\sethlcolor{AdagradColor}\hl{#1}}}

\newmdenv[innerlinewidth=0.5pt, roundcorner=4pt,linecolor=mycolor,innerleftmargin=6pt,
innerrightmargin=6pt,innertopmargin=6pt,innerbottommargin=6pt]{mybox}

\newcommand{\paren}[1]{\left( #1 \right)}
\newcommand{\bm}[1]{
\begin{bmatrix}
#1
\end{bmatrix}
}

\newcommand{\setalglineno}[1]{%
  \setcounter{ALG@line}{\numexpr#1-1}}

\allowdisplaybreaks

\newcommand{\mb}[1]{\mathbf{#1}}
\newcommand{\real}[1]{\mathbb{R}^{#1}}
\newcommand{\minp}[2]{{
\begin{minipage}[t]{#1}
\begin{center}
#2
\end{center}
\end{minipage}
}}
\newcommand{\sets}{\leftarrow}

\DeclareCaptionType{algo}[Algorithm][List of Algorithms]

\title{Addressing Heterogeneity in Federated\\Load Forecasting with Personalization Layers}

\author{
Shourya Bose, Yu Zhang\\Department of Electrical and Computer Engineering\\University of California, Santa Cruz, Santa Cruz, CA \\\\
Kibaek Kim\\Mathematics and Computer Science Division\\Argonne National Laboratory, Lemont, IL}

\begin{document}
\maketitle

\begin{abstract}
{\small The advent of smart meters has enabled pervasive collection of energy consumption data for training short-term load forecasting models. In response to privacy concerns, federated learning (FL) has been proposed as a privacy-preserving approach for training, but the quality of trained models degrades as client data becomes heterogeneous. In this paper we propose the use of personalization layers for load forecasting in a general framework called PL-FL. We show that PL-FL outperforms FL and purely local training, while requiring lower communication bandwidth than FL. This is done through extensive simulations on three different datasets from the NREL ComStock repository.}
\end{abstract}

\section*{Keywords}
Model personalization, federated learning, machine learning, load forecasting, smart grid

\section{Introduction}

Smart grid operations rely on effectively forecasting electrical load demands of different participants in the grid. While commercial smart meters are capable of collecting load data at intervals on the scale of a few minutes, this capability so far has not translated into widespread adoption of residence-level short-term load forecasting models. A principal reason for this is the privacy-sensitive nature of smart meter data~\cite{MRA-etal:2017}, which has led to many legislation against downloading such data to third-party servers. On the other hand, utility operators require high-quality forecasts in order to balance loads in grids with high renewable penetration. This quandary can be resolved with federated learning (FL), wherein a shared forecasting model is trained cooperatively by all smart meters (henceforth called \emph{clients}) and a server, all while ensuring data localization. The latter is achieved by sharing model parameters instead of raw data during training. There exists a vast literature on the implementation of FL in load forecasting from the perspective of system design~\cite{AT-SC:2020}, communications~\cite{ZM-etal:2023}, choice of forecasting model~\cite{YS-XX-2022}, data leakage from trained models~\cite{MAH-etal:2023} and methods to prevent the same~\cite{MAH-etal:2023,DQ-etal:2023}. Two aspects of load forecasting with FL make it distinct from other FL applications. First, it has been observed that FL performance degrades under data heterogeneity~\cite{AA-etal:2022}, i.e. a single shared model may not sufficiently capture the diversity of load shapes and scales for each client. Secondly, any model and training algorithm used must run on limited computational resources and low communication bandwidth, as is typical for the clients~\cite{Micro}.

The aforementioned issues in the context of FL can be solved with \emph{personalization}, wherein each client generates a model which is biased to its own data, while retaining characteristics of the shared model. Personalization has attracted significant research interest in the context of load forecasting, see for e.g.~\cite{FW-etal}. These works can be classified into two categories, i.e. those using \emph{proximal methods} (e.g.~\cite{AT-SC:2020}), and those using \emph{fine tuning} (e.g.~\cite{briggs2021federated}). In the former, a penalty term is added to the loss during client training which discourages large deviations from the server's model, whereas in the latter a shared model trained using conventional FL is further trained with each client's data to generate personalized models. 

Different from such strategies, we propose a novel approach through utilizing \emph{personalization layers} (PLs) to achieve model personalization with a train algorithm called PL-FL. PL-FL involves marking certain model parameters (or layers) as personalized, and other as shared. PLs are updated using only local data, while shared layer parameters participate in the FL update process (see~\cite{arivazhagan2019federated} for an implementation in computer vision domain). There are two major benefits to using PL-FL over proximal or fine-tuning methods.
\begin{enumerate}
    \item Since only shared layer parameters are communicated during updates as opposed to every parameter, the required communication bandwidth is significantly reduced.
    \item As will be shown empirically, PL-FL achieves better performance than conventional FL and purely local training on load forecasting tasks.
\end{enumerate}

In our numerical study, we compare a wide variety of server and client optimization algorithms reported in literature~\cite{reddi2019convergence,mcmahan2017communication,reddi2021adaptive,wu2023faster} for the purpose of training. For testing different algorithms, we use the long-short term memory (LSTM) based load forecasting model introduced in~\cite{kong2017short}. While there exist more sophisticated models comprising attention-based recurrent neural networks~\cite{ijcai2017p366}, their evaluation time-complexity grows as $O(T^2)$ where $T$ is the length of the past sequence used to generate forecasts. This makes them unsuitable for smart meter applications, as compared to LSTM.

\section{Classical Federated Learning}

Consider clients $c=1,\dots,N$ which want to collectively learn a model $\mb{y}=f_{\pmb{\theta}}(\mb{x})$, where $\pmb{\theta}\in\real{p}$ is the learnable parameter. Each possesses a dataset $D_c=\{\mb{x}_i,\mb{y}_i\}_{i=1}^{n_c}$ and can share model updates to a central server indexed by $c=0$. The basic framework of FL is sketched in Algorithm~\ref{alg:classical}. Herein, the server executes $T_s$ global epochs, each of which involves two communications and a client update. The first communication broadcasts the server's parameters $\pmb{\theta}^c_{t-1}$ to each client. This is followed by each client $c$ updating their local models with \textsc{ClientOpt} for $T_c$ local epochs using their data $D_c$, thereby generating a gradient $\mb{g}_t^c$ representing the local model's update. These gradients are transferred back to the server, which aggregates them in \textsc{ServerOpt} to update the server state $\mb{S}_t$, which contains the server parameters $\pmb{\theta}^0_t$ alongside other elements.

In this paper, we consider an extensive list of federated optimization algorithms as benchmarks for both \textsc{ServerOpt} and \textsc{ClientOpt}. The role of the client optimizers is to minimize a loss function of the local parameters defined at each client's data, with some examples being stochastic gradient descent (SGD), AdaGrad, Adam, etc. We consider four candidates for \textsc{ClientOpt} \emph{viz.} Adam, AdamAMS, Prox, and ProxAdam (see Appendix for details). All but Prox are based on Adam, which is known to produce good training performance due to its adaptivity. Prox and ProxAdam are the proximal variants of SGD and Adam, respectively, to achieve model personalization in FL, as in Section~\ref{sec:CS}. We now move to \textsc{ServerOpt}, which aims to aggregate the gradient of client updates in order to update the server's parameters $\pmb{\theta}^0_t$. We consider FedAvg~\cite{mcmahan2017communication}, FedAvgAdaptive~\cite{wu2023faster}, FedAdagrad~\cite{reddi2021adaptive}, FedAdam~\cite{reddi2021adaptive}, and FedYogi~\cite{reddi2021adaptive} as candidates. All these algorithms maintain a single set of parameters except FedAvgAdaptive, which maintains distinct parameters for each client, thereby providing a degree of personalization from the outset.
As mentioned earlier, both server and client states contain additional elements which aids the optimization and aggregation. For example, adaptive algorithms maintain states $\mb{m}$ and $\mb{v}$ apart from model parameters, which are used to track the exponential moving average of the gradients and their second moments. This is the case for all server and client algorithms we consider except FedAvg and Prox.\\

\begin{algorithm}[!tb]
\begin{algorithmic}[1]
\STATE Initialize server state $\mb{S}_0$\label{alg:classical}
\STATE Locally initialize client states $\mb{C}^c_0$ for $c=1$ to $N$
\FOR{$t=1$ to $T_s$}
    \FOR{$c=1$ to $N$}
        \STATE Extract $\pmb{\theta}_{t-1}^0$ from $\mb{S}_{t-1}$
        \STATE $\textsc{Communicate}(\pmb{\theta}_{t-1}^0, 0\rightarrow c)$
        \STATE $\mb{C}^c_t,\mb{g}_t^c \leftarrow \textsc{ClientOpt}(\pmb{\theta}_{t-1}^0,\mb{C}^c_{t-1},D_c,T_c)$
        \STATE $\textsc{Communicate}(\mb{g}_t^c, c\rightarrow 0)$
    \ENDFOR
    \STATE $\mb{S}_t \leftarrow \textsc{ServerOpt}(\mb{S}_{t-1},\mb{g}_t^1,\cdots,\mb{g}_t^N)$
\ENDFOR
\STATE from $\mb{S}_{T_s}$, extract $\pmb{\theta}^0_{Ts}$ and \textbf{return} $\pmb{\theta}^0_{T_s}$
\end{algorithmic}
\caption{Classical FL}
\end{algorithm}
\begin{algorithm}[!tb]
\caption{Personalization Layers for FL (PL-FL)}
\label{alg:algorithm_label}
\begin{algorithmic}[1]
\STATE Initialize shared server state $\mb{S}_0^{sh}$
\STATE Locally initialize client states $\mb{C}^c_0$ for $c=1$ to $N$\label{alg:pers}
\FOR{$t=1$ to $T_s$}
    \FOR{$c=1$ to $N$}
        \STATE Extract $\pmb{\theta}_{t-1}^{0,sh}$ from $\mb{S}_{t-1}^{sh}$
        \STATE $\textsc{Communicate}(\pmb{\theta}_{t-1}^{0,sh}, 0\rightarrow c)$
        \STATE Extract $\pmb{\theta}_{t-1}^{0,pe}$ from $\mb{C}^c_{t-1}$
        \STATE $\pmb{\theta}^0_{t-1}\sets[\pmb{\theta}_{t-1}^{0,sh},\pmb{\theta}_{t-1}^{0,pe}]$
        \STATE $\mb{C}^c_t,\mb{g}_t^c \leftarrow \textsc{ClientOpt}(\pmb{\theta}_{t-1}^0,\mb{C}^c_{t-1},D_c,T_c)$
        \STATE Extract $\mb{g}^{c,sh}_t$ from $\mb{g}^c_t$
        \STATE $\textsc{Communicate}(\mb{g}_t^{c,sh}, c\rightarrow 0)$
    \ENDFOR
    \STATE $\mb{S}_t^{sh} \leftarrow \textsc{ServerOpt}(\mb{S}_{t-1}^{sh},\mb{g}_t^{1,sh},\cdots,\mb{g}_t^{N,sh})$
\ENDFOR
\STATE from $\mb{S}_{T_s}$, extract $\pmb{\theta}^0_{Ts}$ and \textbf{return} $\pmb{\theta}^0_{T_s}$
\end{algorithmic}
\end{algorithm}

\section{Personalization Layers for FL (PL-FL)}

Two major factors motivate the need for PL-FL. Firstly, we want to tackle the non-i.i.d (independent and identically distributed) nature of data at each client, for which personalization can be a possible solution. Secondly, we see in Algorithm~\ref{alg:classical} that on every global epoch, the communication of vectors $\pmb{\theta}_t^0$ and $\mb{g}^c_t$ between each client and the server is required. While it is possible to employ compression schemes to reduce the size of communications, personalization allows for the transfer of smaller vectors pertaining only to shared layers, thereby providing a principled yet effective means of communication reduction. Note that for proximal methods as implemented in~\cite{AT-SC:2020}, the second problem remains: full parameter and gradient vectors are communicated on every epoch.

We introduce PL-FL for the purpose of resolving this issue, wherein each parameter (or possibly layer) is marked either \emph{shared} (denoted with superscript \emph{sh}) and \emph{personalized} (denoted with superscript \emph{pe}). In other words, we split the model parameters as $\pmb{\theta}^c_t=[\pmb{\theta}^{c,sh}_t,\pmb{\theta}^{c,pe}_t]$ (similarly for any associated states such as $\mb{m}$ and $\mb{v}$). Different from classical FL, PL-FL only communicates shared parameters with the server, while letting personalized ones remain local to each client. Since most server algorithms (including the ones we consider in this paper) carry out parameter-wise aggregation, it is possible to personalize any arbitrary subset of parameters. The algorithmic steps of PL-FL are described in Algorithm~\ref{alg:pers}. We summarize three potential benefit of PL-FL.
\begin{itemize}
    \item \textbf{Communication Efficiency:} The amount of communication needed reduces since we transfer $(\pmb{\theta}^{sh}_{t-1},\mb{g}^{c,sh}_{t-1})$ as opposed to the larger $(\pmb{\theta}_{t-1},\mb{g}^{c}_{t-1})$.
    \item \textbf{Combatting Heterogeneity:} The effect of data heterogeneity (non-i.i.d data) on FL is mitigated by letting parameters $\pmb{\theta}^{c,pe}_t$ remain local to each client. This allows better fitting of local data, while a common shared representation is learned via shared parameters $\pmb{\theta}^{c,sh}_t$.
    \item \textbf{Model Privacy:} Since the server never possesses a full set of parameters, inference remains local to each client, thereby enhancing privacy. Third parties may request forecasts from local models as needed, which are computed locally.
\end{itemize}

As compared to PL-FL, proximal methods do not offer the first benefit since the server has to communicate the full weights $\pmb{\theta}^0_t$ on every global epoch. Furthermore, we show in Section~\ref{sec:CS} that PL-FL offers much better performance than using purely proximal methods. On the other hand, proximal methods themselves can be made amenable to PL-FL by only considering proximal penalties corresponding to shared parameters which we test. Furthermore, we forgo testing more complex server algorithms such as FedDANE~\cite{li2019feddane} and SCAFFOLD~\cite{karimireddy2020scaffold} due to multiple back-and-forth communications needed per server epoch, which increases computation and communication overheads.

\section{Case Study}
\label{sec:CS}

In this section, we compare different algorithms using the LSTM load forecasting model and NREL ComStock dataset~\cite{parker2023comstock}. The trained model accuracy on each client's model, as well as the amount of communication are compared.

\subsection{Load Forecasting Model}
\begin{algorithm}[!tbh]
\caption{LSTMCell\\ \emph{$\sigma\paren{\, \cdot \,}$ denotes the sigmoid function while $\odot$ represents elementwise product}}
\label{alg:algorithm_label}
\begin{algorithmic}[1]
\STATEx \textbf{Input:} $\mb{h}_{t-1},\mb{c}_{t-1},\mb{x}_{t}$
\STATEx \textbf{Parameters:} $\mb{W}_{ii},\mb{W}_{hi},\mb{b}_i,\mb{W}_{if},\mb{W}_{hf},\mb{b}_f,$
\STATEx $\mb{W}_{ig},\mb{W}_{hg}, \mb{b}_g,\mb{W}_{io},\mb{W}_{ho},\mb{b}_o$
\STATE $\mb{i}_t \sets \sigma\paren{\mb{W}_{ii}\mb{x}_t + \mb{W}_{hi}\mb{h}_{t-1}+\mb{b}_i}$
\STATE $\mb{f}_t \sets \sigma \paren{\mb{W}_{if}\mb{x}_t + \mb{W}_{hf}\mb{h}_{t-1}+\mb{b}_f}$
\STATE $\mb{g}_t \sets \text{tanh}\paren{\mb{w}_{ig}\mb{x}_t+\mb{W}_{hg}\mb{h}_{t-1}+\mb{b}_g}$\label{alg:lstm}
\STATE $\mb{o}_t = \sigma \paren{\mb{W}_{io}\mb{x}_t + \mb{W}_{ho}\mb{h}_{t-1}+\mb{b}_o}$
\STATE $\mb{c}_t = \mb{f}_t \odot \mb{c}_{t-1} + \mb{i}_t \odot \mb{g}_t$
\STATE $\mb{h}_t = \mb{o}_t \odot \text{tanh}\paren{\mb{c}_t}$
\STATE return $\mb{h}_t,\mb{c}_t$
\end{algorithmic}
\end{algorithm}

The LSTM model is capable of learning load forecasting problems of the form $\hat{y}_{T+L}=f_{\pmb{\theta}}(y_1,\cdots,y_T,\mb{z}_1,\cdots,\mb{z}_T)$. Here $y_t\in\real{}$ represents the load, while $\mb{z}_t\in\real{d}$ represents additional features such as datetime indices, weather forecasts, building characteristics, etc. The quantities $T$ and $L$ are called look-back and look-ahead, and denote the number of past data points ($T$) required to estimate the demand $L$ steps in the future. Therefore, the inputs $\mb{x}_i$ and labels $\mb{y}_i$ of any dataset over which LSTM trains are given as
\begin{gather*}
    \mb{x}_i = \left\{ \bm{y_1\\ \mb{z}_1},\cdots,\bm{y_T\\ \mb{z}_T}\right\},\quad \mb{y}_i = y_{T+L},
\end{gather*}
wherein each input-label pair $i$ may be drawn from any random point in the historical data stored at each client. LSTM at its core consists of a recursive rollout of a LSTMCell kernel (Algorithm~\ref{alg:lstm}), which ingests previous hidden and cell states $\mb{h}_{t-1},\mb{c}_{t-1}$ as well as the current input $\mb{x}_t$ to generate updated states $\mb{h}_{t},\mb{c}_{t}$. The final forecast $\hat{y}_{T+L}$ is generated by passing the hidden states $\mb{h}_1,\cdots,\mb{h}_{T}$ through a multilayer perceptron (MLP) to generate the forecast $\hat{y}_{T+L}$. Overall, LSTM inference can be described as
\begin{gather*}
    \mb{h}_{t},\mb{c}_t = \text{LSTMCell} (\mb{h}_{t-1},\mb{c}_{t-1},\mb{x}_t), \forall t \in \{1,\cdots,T\}\\
    \mb{h}_0=\mb{c}_0 = \mb{0},\quad \hat{y}_{T+L} = \text{MLP}(\mb{h}_1,\cdots,\mb{h}_T).
\end{gather*}
Note that the total parameter $\pmb{\theta}$ of the LSTM model includes parameters from the LSTMCell as well as those of the MLP. For the purpose of PL-FL, we consider three configurations:
\begin{itemize}
    \item \textbf{P1:} All parameters shared (i.e. FL)
    \item \textbf{P2:} LSTMCell parameters shared, MLP parameters personalized
    \item \textbf{P3:} All parameters personalized (i.e. purely local training)
\end{itemize}
Comparisons between these three schemes enable us to evaluate the benefits of PL-FL. In order to keep the model size small and amenable to running on clients, we choose $\mb{h}_t,\mb{c}_t\in\real{25}$ and use a MLP with layer input sizes $(300,150,75)$ and PReLU activations.

\subsection{Dataset} 
\begin{figure}[H]
    \centering
    \includegraphics[width=\linewidth]{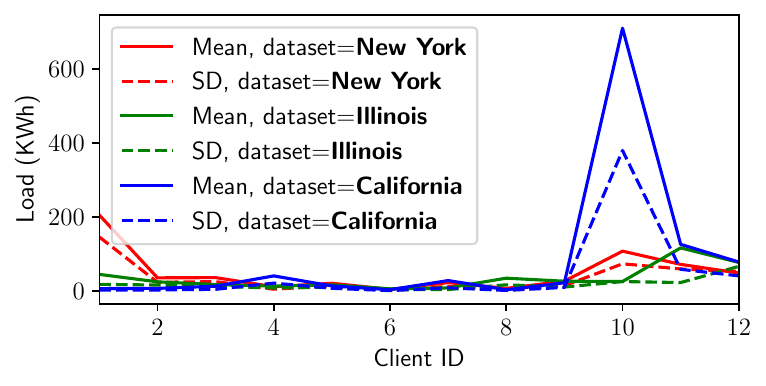}
    \caption{Mean and standard deviation of loads of 12 clients each from the tree datasets.}
    \label{fig:var}
\end{figure}
\begin{figure}[H]
    \centering
    \includegraphics[width=\linewidth]{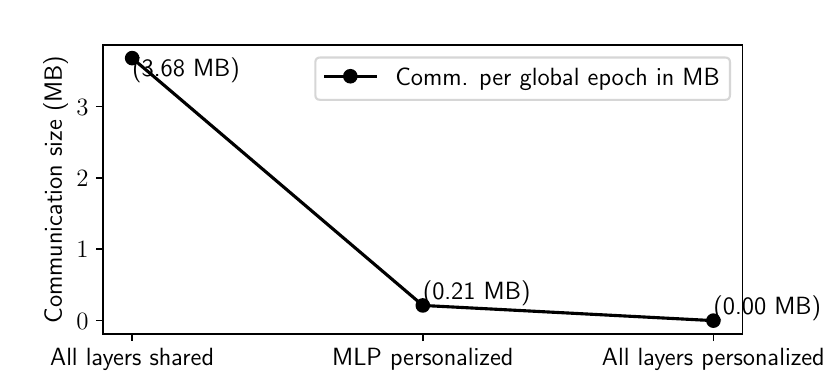}
    \caption{Total data transferred per global epoch per client. This consists of a bidirectional communication between the server and client.}
    \label{fig:COMM}
\end{figure}

We use the NREL ComStock dataset, which is a semi-synthetic dataset consisting of load shapes and other features for different commercial buildings across all 50 US states. We choose 12 randomly selected buildings (corresponding to clients) from the states of California, Illinois and New York, which are treated as three separate experiments. Apart from loads, we select $d=7$ features for $\mb{z}_t$, which include date and time indices, temperature, windspeed, and static building features such as floor space, wall area, and window area.
The heterogeneity in clients' load data is reflected in Figure~\ref{fig:var}. Herein, it can be seen that the temporal mean and standard deviation of each client's loads differ significantly in scale, which reflects the reality of different buildings having diverse load consumption patterns.

The dataset contains one year's data at a granularity of 15 minutes. We split the data along the time axis in a ratio of $0.8:0.1:0.1$ as training, validation, and test sets respectively. We choose a look-back of $T=12$ (3 hours) to generate a forecast $L=4$ steps (1 hour) into the future. In the interest of reducing computational burden, we choose a small batch size of 16.

\subsection{Setup and Simulation Results}
\begin{figure*}[ht!]
    \centering
    \includegraphics[width=0.83\linewidth]{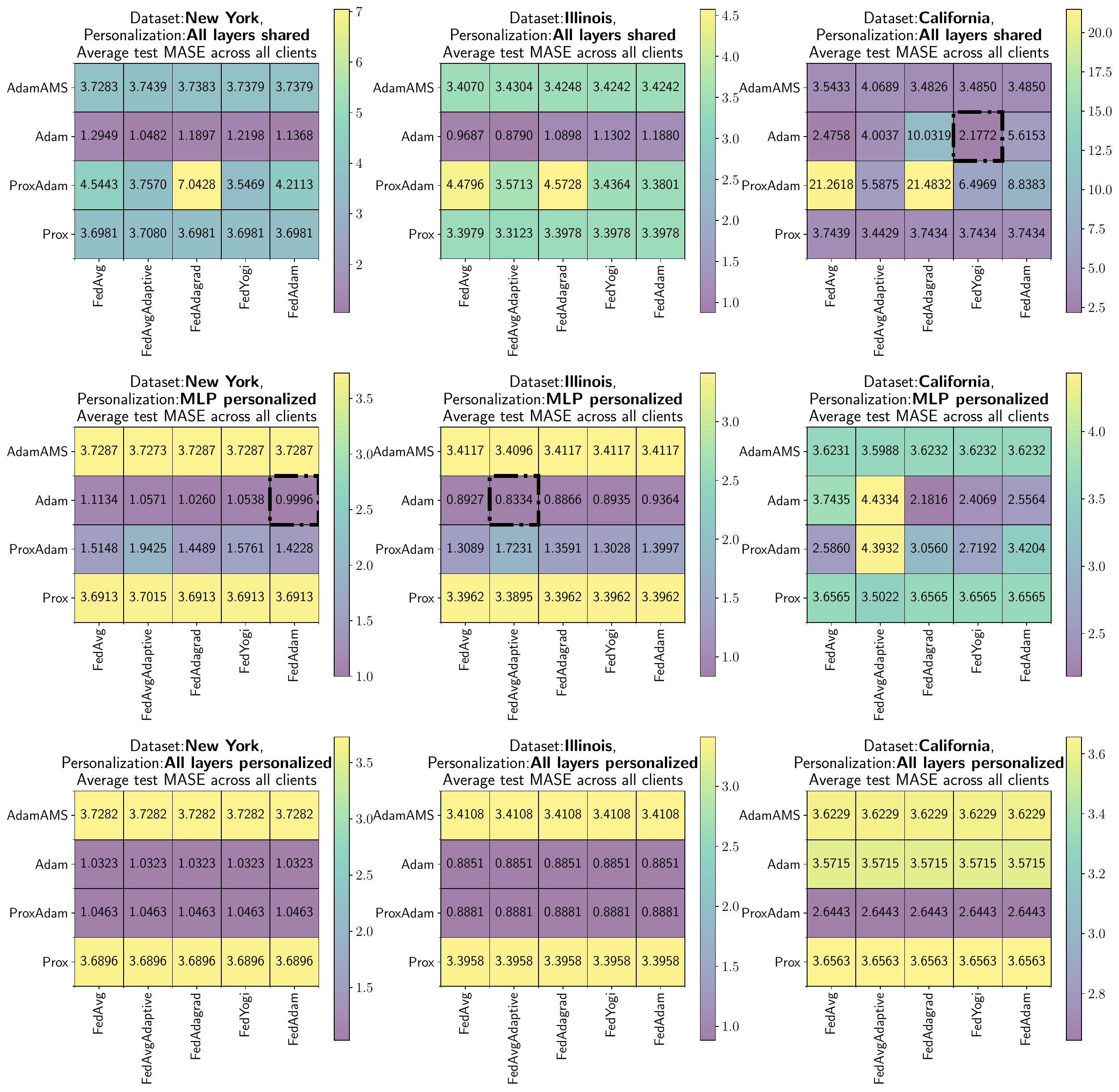}
    \caption{Average MASE metric on the test set for different client and server algorithms via PL-FL. These are calculated for local training, MLP personalization, and all layers shared (FL). Three datasets \emph{viz.} New York, Illinois, and California are used.}
    \label{fig:MASE}
\end{figure*}
All code is written in PyTorch with the aid of the APPFL package~\cite{APPFL}, and implemented using PyTorch and OpenMPI. We use MASE errors,
\begin{align*}
    \text{MASE} = \frac{ \sum_{t} |\hat{x}_t - x_t |}{\sum_t|x_{t-L} - x_t|},
\end{align*}
wherein it is assumed that both sequences are aligned and the number of terms in the numerator and denominator are equal. MASE values of less than 1 can be interpreted as the forecasts being of a better quality than the persistent method, i.e. forecasting the last known data point. The results can be seen in Figure~\ref{fig:MASE}. Immediately, it is obvious that adaptive methods such as Adam and AdamAMS perform better than proximal methods across all tests. Furthermore, for all three datasets, the best performance is achieved when the MLP is personalized and Adam is used, thereby showing the advantages of local adaptive optimization. Combined with the per-epoch data transfer shown in Figure~\ref{fig:COMM}, it is obvious that PL-FL with adaptive algorithms performs better than even purely local training, while taking a fraction of the bandwidth of FL. However, it is important to note that none of the tests achieved a MASE of $<1$ on the California dataset, which might indicate the need for more sophisticated forecast models or better feature selection for that dataset.
\section{Conclusion}
In this paper, we introduced PL-FL, which is a framework to train personalized load forecasting models for smart meter clients. We tested a variety of client and server algorithms over multiple datasets, which showed that PLs are effective in improving FL across all settings. Future work involves the design of distributed model inference schemes, which enables practical implementation of PL-FL on commercial smart meter hardware.



\begin{figure*}
\label{sec:Appendix}
\appendix
\section{Appendix}
    \centering
    \newcommand{\param}{$\{ \pmb{\theta}^c_t,\mb{m}^c_t,\mb{v}^c_t\}$}
    \begin{tabular}{c|c|c}
        \textbf{Optimizer} & \textbf{Local State $\mb{C}^c_t$} & \textsc{ClientOpt} \\
        \hline
        \minp{1.75cm}{\hladam{Adam}\\\hlams{AdamAMS}\\\hlprox{Prox}\\\hlpadam{ProxAdam}} & \minp{1.75cm}{\hladam{\param}\\\hlams{\param}\\\hlprox{$\{\pmb{\theta}^c_t\}$}\\\hlpadam{\param}} & \minp{10cm}{
            \begin{algorithmic}[1]
                \STATEx \textbf{Inputs: $\pmb{\theta}^0_{t-1},\mb{C}^c_{t-1},D_c,T_c$}
                \STATEx \textbf{Hyperparameters:} Client learning rate $\eta$, Adaptivity parameters $\beta_1,\beta_2,\epsilon$ (for \hladam{Adam},\hlams{AdamAMS},\hlpadam{ProxAdam})
                \STATE Extract $\pmb{\theta}^c_{t-1}$ from $\mb{C}^c_{t-1}$ and set $\pmb{\theta}^c_{t,0}\leftarrow\pmb{\theta}^c_{t-1}$ \COMMENT{\hlams{AdamAMS},\hlpadam{ProxAdam}}
                \STATE Initialize $\mb{m}^c_{t,0}$, $\mb{v}^c_{t,0}, \mb{v}^{c,\max}_{t,0}$ depending on optimizer \label{line:localInit}
                \FOR{$t'=1$ to $T_c$}
                    \STATE Sample minibatch $D_{c,t'}$ from $D_c$ of size $B$
                    \STATE $\mb{g}_{t,t'-1}^c \leftarrow\frac{1}{B} \sum\limits_{((\mb{x},\mb{y})\in D_{c,t'}} \hskip -0.5cm \nabla_{\pmb{\theta}}(f_{\pmb{\theta}}(\mb{x}) - \mb{y})^2 $ \COMMENT{{\hladam{Adam},\hlams{AdamAMS}}}
                    \setalglineno{6}
                    \STATE $\mb{g}_{t,t'-1}^c \leftarrow\frac{1}{B} \sum\limits_{((\mb{x},\mb{y})\in D_{c,t'}} \hskip -0.5cm \nabla_{\pmb{\theta}}\left( (f_{\pmb{\theta}}(\mb{x}) - \mb{y})^2   + \alpha\lVert \pmb{\theta}-\pmb{\theta}^0_{t-1}\rVert_2^2 \right) \big\vert_{\pmb{\theta}\sets\pmb{\theta}^c_{t,t'-1}}$ \COMMENT{{\hlprox{Prox}, \hlpadam{ProxAdam}}}
                    \STATE $\mb{m}^c_t \sets \beta_1\mb{m}^c_{t-1}+(1-\beta_1)\mb{g}_{t,t'-1}^c$ \COMMENT{\hladam{Adam},\hlams{AdamAMS},\hlpadam{ProxAdam}}
                    \STATE $\mb{v}^c_{t,t'} \sets \beta_2\mb{v}^c_{t-1}+(1-\beta_2)(\mb{g}_{t,t'-1}^c)^2$ \COMMENT{\hladam{Adam},\hlams{AdamAMS},\hlpadam{ProxAdam}}
                    \STATE $\hat{\mb{m}}^c_{t,t'},\hat{\mb{v}}^c_{t,t} \sets \frac{\mb{m}^c_{t,t'}}{1-\beta_1^t}, \frac{\mb{v}^c_{t,t'}}{1-\beta_2^t}$ \COMMENT{\hladam{Adam},\hlams{AdamAMS},\hlpadam{ProxAdam}}
                    \STATE $\mb{v}^{c,\max}_{t,t'} \sets \max\paren{\mb{v}^{c,\max}_{t,t'-1},\hat{\mb{v}}^c_{t,t'}}$ \COMMENT{\hlams{AdamAMS}}
                    \STATE $\pmb{\theta}^c_{t,t'} \leftarrow \pmb{\theta}^c_{t,t'-1} - \eta\mb{g}^c_{t,t'-1}$ \COMMENT{\hlprox{Prox}}
                    \setalglineno{11}
                    \STATE $\pmb{\theta}^c_{t,t'} \leftarrow \pmb{\theta}^c_{t,t'-1} - \eta\frac{\hat{\mb{m}}^c_{t,T_c}}{\sqrt{\hat{\mb{v}}^c_{t,t'}}+\epsilon}$ \COMMENT{\hladam{Adam},\hlpadam{ProxAdam}}
                    \setalglineno{11}
                    \STATE $\pmb{\theta}^c_{t,t'} \leftarrow \pmb{\theta}^c_{t,t'-1} - \eta\frac{\hat{\mb{m}}^c_{t,t'}}{\sqrt{\hat{\mb{v}}^{c,\max}_{t,T_c}}+\epsilon}$ \COMMENT{\hlams{AdamAMS}}
                \ENDFOR
                \STATE $\mb{g}^c_t \sets \pmb{\theta}^c_{t,0} - \pmb{\theta}^c_{t,T_c}$
                \STATE $\mb{C}^c_t \sets \{ \theta^c_t = \theta^c_{t,T_c}\}$\COMMENT{\hlprox{Prox}}
                \setalglineno{14}
                \STATE $\mb{C}^c_t \sets \{ \theta^c_t = \theta^c_{t,T_c},\mb{m}^c_t=\mb{m}^c_{t,T_c},\mb{v}^c_t=\mb{v}^c_{t,T_c}\}$\COMMENT{\hladam{Adam},\hlams{AdamAMS},\hlpadam{ProxAdam}}
                \STATE \textbf{return} $\mb{C}^c_t$, $\mb{g}^c_t$
            \end{algorithmic}
        }\\
        \hline
    \end{tabular}
    \captionof{figure}{A schematic of \textsc{ClientOpt} for different client algorithms. Note that in line~\ref{line:localInit}, client states $\mb{m}$ or $\mb{v}$ are reinitialized rather than inheriting stale values to ensure better performance. All vector operations here are elementwise.}
    \label{fig:LOCAL}
    \end{figure*}
    \begin{figure*}
    \centering
    \newcommand{\param}{$\{ \pmb{\theta}^c_t,\mb{m}^c_t,\mb{v}^c_t\}$}
    \begin{tabular}{c|c|c}
        \textbf{Optimizer} & \textbf{Server State $\mb{S}_t$} & \textsc{ServerOpt} \\
        \hline
        \minp{2.5cm}{\hlfav{FedAvg}\\ \hlfada{FedAdagrad}\\ \hlfyogi{FedYogi}\\ \hlfadam{FedAdam}\\ \hlfadapt{FedAvgAdaptive}} & \minp{1.5cm}{\hlfav{$\{\pmb{\theta}_t^0\}$}\\\hlfada{$\{\pmb{\theta}_t^0,\mb{m}^0_t,\mb{v}^0_t \}$} \\ \hlfyogi{$\{\pmb{\theta}_t^0,\mb{m}^0_t,\mb{v}^0_t \}$} \\ \hlfadam{$\{\pmb{\theta}_t^0,\mb{m}^0_t,\mb{v}^0_t \}$} \\ \hlfadapt{$\{\{\pmb{\theta}_t^{0,(c)},$\\$\mb{v}^{0,(c)}_t\}_{c=1}^N \}$}} & \minp{10cm}{
            \begin{algorithmic}[1]
                \STATEx \textbf{Inputs: $\pmb{\theta}^0_{t-1},\mb{C}^c_{t-1},D_c,T_c$}
                \STATEx \textbf{Hyperparameters:} Server learning rate $\eta_s$, Server adaptivity parameters $\beta_{1s},\beta_{2s},\epsilon_s$ (for \hlfada{FedAdagrad}, \hlfyogi{FedYogi}, \hlfadam{FedAdam}, \hlfadapt{FedAvgAdaptive})
                \STATE $\bar{\mb{g}}_t \sets \sum_{c=1}^N \mb{\mb{g}}^c_t$
                \STATE $\mb{m}^0_t \sets \beta_{1s}\mb{m}^0_{t-1}+(1-\beta_{1s})\bar{\mb{g}}_t$ 
                \COMMENT{\hlfada{FedAdagrad},\hlfyogi{FedYogi},\hlfadam{FedAdam}}
                \STATE $\mb{v}^0_t \sets \paren{\bar{\mb{g}}_t}^2$
                \COMMENT{\hlfada{FedAdagrad}}
                \setalglineno{3}
                \STATE $\mb{v}^0_t \sets \beta_{2s}\mb{v}^0_{t-1} - (1-\beta_{2s})\paren{\bar{g}_t}^2\text{sign}\paren{\mb{v}^0_{t-1}-\paren{\bar{g}_t}^2}$
                \COMMENT{\hlfyogi{FedYogi}}
                \setalglineno{3}
                \STATE $\mb{v}^0_t \sets \beta_{2s}\mb{v}^0_{t-1} - (1-\beta_{2s})\paren{\bar{g}_t}^2$
                \COMMENT{\hlfadam{FedAdam}}
                \STATE $\pmb{\theta}^0_t \sets \pmb{\theta}^0_{t-1} + \eta_s \bar{\mb{g}}_t$
                \COMMENT{\hlfav{FedAvg}}
                \STATE $\pmb{\theta}^0_t \sets \pmb{\theta}^0_{t-1} + \eta_s \frac{\mb{m}^0_t}{\sqrt{\mb{v}^0_t}+\epsilon_s}$ \COMMENT{\hlfada{FedAdagrad},\hlfyogi{FedYogi},\hlfadam{FedAdam}}
                \FOR{$c =1$ to $N$}
                    \STATE $\mb{v}^{0,(c)}_t \sets \beta_{1s}\mb{v}^{0,(c)}_{t-1}+(1-\beta_{1s})\paren{\mb{g}^c_t}^2$ \COMMENT{\hlfadapt{FedAvgAdaptive}}
                    \STATE $\pmb{\theta}^{0,(c)}_{t} \sets \sum_{c=1}^C \paren{ \pmb{\theta}^{0,(c)}_{t-1} -\eta_s \frac{ \mb{g}^c_t }{\sqrt{\mb{v}^{0,(c)}}+\epsilon_s}}$ \COMMENT{\hlfadapt{FedAvgAdaptive}}
                \ENDFOR
                \STATE Populate state $\mb{S}_t$ and \textbf{return}
            \end{algorithmic}
        }\\
        \hline
    \end{tabular}
    \caption{A schematic of \textsc{ServerOpt} for different server algorithms. All algorithms use a single state except FedAvgAdaptive, which uses $N$ distinct states for each of the clients. All vector operations here are elementwise.}
    \label{fig:GLOBAL}
\end{figure*}
\end{document}